\documentclass{article}
\usepackage[T1]{fontenc}     
\usepackage{ragged2e}        
\usepackage[protrusion=true,expansion=false]{microtype}  
\usepackage{graphicx}
\usepackage{multirow}
\usepackage{caption}
\usepackage{float}
\usepackage{booktabs}
\usepackage{multicol}
\usepackage{tablefootnote}
\usepackage{textgreek} 
\usepackage{tabularx}        
\usepackage[table,xcdraw]{xcolor}
\usepackage{amsmath,amssymb,amsfonts}
\usepackage{mathtools} 
\usepackage{amsthm}
\usepackage{mathrsfs}
\usepackage{algorithm}
\usepackage{algpseudocode}  
\usepackage{algcompatible}%
\usepackage{algorithmicx}

\usepackage{textcomp}
\usepackage{pifont}
\usepackage{setspace,lineno}
\usepackage{listings}
\usepackage{soul}            
\sethlcolor{yellow}          
\usepackage[title]{appendix}
\usepackage{manyfoot}
\usepackage{xcolor}

\definecolor{customgreen}{HTML}{00B050}
\definecolor{captionblue}{HTML}{0070C0}  

\DeclareCaptionLabelFormat{boldblue}{\textbf{\textcolor{captionblue}{#1 #2}}}
\usepackage{titlesec}
\titleformat{\section}
  {\normalfont\Large\bfseries\color{captionblue}}{\thesection}{1em}{}
\titleformat{\subsection}
  {\normalfont\large\bfseries\color{captionblue}}{\thesubsection}{1em}{}
\titleformat{\subsubsection}
  {\normalfont\normalsize\bfseries\color{captionblue}}{\thesubsubsection}{1em}{}
\usepackage{geometry}
\usepackage{amsmath} 
\allowdisplaybreaks[4]
\usepackage[protrusion=true,expansion=false]{microtype}
\usepackage{hyperref}
\hypersetup{
    colorlinks=true,
    linkcolor=blue,
    citecolor=blue,
    urlcolor=blue
}

\title{PPORLD-EDNetLDCT: A Proximal Policy Optimization-Based Reinforcement Learning Framework for Adaptive Low-Dose CT Denoising}

\author{
Debopom Sutradhar\textsuperscript{1}, 
Ripon Kumar Debnath\textsuperscript{1},
Mohaimenul Azam Khan Raiaan\textsuperscript{1},\\
Yan Zhang\textsuperscript{2}, 
Reem E. Mohamed\textsuperscript{3}, 
Sami Azam\textsuperscript{2,*}\\
\small
\textsuperscript{1} Department of Computer Science and Engineering, United International University, Dhaka 1212, Bangladesh \\
\small
\textsuperscript{2}Faculty of Science and Technology, Charles Darwin University, Casuarina, NT 0909, Australia\\
\small
\textsuperscript{3}Faculty of Science and Information Technology, Charles Darwin University, Sydney, NSW, Australia\\
\small 
\textsuperscript{*}Corresponding Author: \href{mailto:sami.azam@cdu.edu.au}{sami.azam@cdu.edu.au}
}

\date{} 
\begin{document}


\justifying
\twocolumn[
\maketitle
\begin{abstract}  
\noindent Low-dose computed tomography (LDCT) is critical for minimizing radiation exposure, but it often leads to increased noise and reduced image quality. Traditional denoising methods, such as iterative optimization or supervised learning, often fail to preserve image quality. To address these challenges, we introduce PPORLD-EDNetLDCT, a reinforcement learning-based (RL) approach with Encoder-Decoder for LDCT. Our method utilizes a dynamic RL-based approach in which an advanced posterior policy optimization (PPO) algorithm is used to optimize denoising policies in real time, based on image quality feedback, trained via a custom gym environment. The experimental results on the low dose CT image and projection dataset demonstrate that the proposed PPORLD-EDNetLDCT model outperforms traditional denoising techniques and other DL-based methods, achieving a peak signal-to-noise ratio of 41.87, a structural similarity index measure of 0.9814 and a root mean squared error of 0.00236. Moreover, in NIH-AAPM-Mayo Clinic Low Dose CT Challenge dataset our method achieved a PSNR of 41.52, SSIM of 0.9723 and RMSE of 0.0051. Furthermore, we validated the quality of denoising using a classification task in the COVID-19 LDCT dataset, where the images processed by our method improved the classification accuracy to 94\%,  achieving 4\% higher accuracy compared to denoising without RL-based denoising.
\end{abstract}

\vspace{0.5em}
\vspace{1em}
]

\section{Introduction}
\label{intro}
Computed tomography (CT) is a widely used imaging technique that provides detailed cross-sectional views of the human body\cite{islam2023introduction}. It is essential for the detection of various disorders and the development of treatment plans \cite{hussain2022modern}. However, CT imaging relies on ionizing radiation, which poses significant safety concerns, especially for people requiring multiple scans, as repeated exposure can increase the risk of adverse health effects \cite{hussain2022modern}. This risk requires a careful balance between obtaining high-quality diagnostic images and minimizing radiation exposure \cite{adelodun2024comprehensive}. Over the years, the benefits of CT have been tempered by the potential harm of radiation, prompting ongoing research toward safer imaging protocols \cite{dudhe2024radiation}. Addressing these challenges is crucial for improving patient care while maintaining the diagnostic capabilities of CT scans \cite{huang2021gan}.

Low-dose CT (LDCT) imaging was developed to reduce the risks associated with radiation exposure \cite{zubair2024enabling}. By lowering the radiation dose, LDCT offers a safer alternative for patients, particularly those who need frequent scans to monitor or detect early disease \cite{yip2021added}. This approach helps in screening programs and routine examinations where reducing radiation is a priority, yet it still provides clinically useful images \cite{nanni2024eanm}. It has become a valuable tool in modern diagnostics, as it minimizes the potential harm without completely compromising the ability to detect critical conditions \cite{nanni2024eanm}. Its implementation has improved patient safety and broadened the use of CT imaging in preventive care while limiting exposure risks \cite{huang2021gan}.

Despite its benefits, LDCT imaging introduces new challenges. For example, a reduced radiation dose often results in images with increased noise and artifacts, which can obscure important anatomical details and compromise diagnostic accuracy. The degraded image quality in LDCT scans significantly impacts downstream diagnostic tasks, including automated disease classification and clinical decision-making processes, requiring effective denoising solutions that preserve diagnostically relevant features. Traditional denoising methods, such as sinogram-based techniques using bilateral and adaptive filtering, tend to blur fine details and miss important edge information \cite{patwari2020jbfnet} \cite{yang2022low}. Iterative reconstruction methods, although effective in some cases, require significant computing resources, making them less suitable for everyday clinical use. Post-processing methods, including non-local means and block matching algorithms (BM3D) \cite{zhang2021r3l}, also struggle to completely eliminate irregular noise and residual artifacts \cite{patwari2020jbfnet} \cite{bai2021probabilistic}. Therefore, the development of effective denoising techniques is critical to improve image quality while preserving essential diagnostic features in LDCT scans.

Recently, deep learning methods have achieved significant success in LDCT denoising, using CNNs, GANs, hybrid architectures, and transformer-based models \cite{huang2021gan} \cite{yang2022low} \cite{tan2022selective} \cite{gu2021adain} \cite{li2021low}\cite{li2023multi}\cite{li2024unpaired}\cite{zhang2023structure} \cite{liu2025diffusion} \cite{jung2022patch} \cite{chen2023ascon} \cite{zhang2021task} \cite{jin2025predictions} \cite{jin2025peanut}. CNN-based methods such as RED-CNN effectively transform low-dose images into near normal-dose quality and achieve notable noise reduction \cite{tan2022selective}. GAN variants, including CycleGAN and DU-GAN, preserve texture and structural details, although they are prone to training instability and mode collapse \cite{huang2021gan} \cite{gu2021adain} \cite{li2021low}. Hybrid approaches that integrate attention mechanisms, multiscale feature fusion, and wavelet transforms -such as MSFLNet \cite{li2023multi}, PCCNN \cite{li2024unpaired}, and AGC-LSRED \cite{zhang2023structure} - excel in maintaining fine details while minimizing artifacts. More recently, transformer-based solutions such as SIST \cite{yang2022low}, zero shot diffusion models\cite{liu2025diffusion}, patchwise deep metric learning \cite{jung2022patch}, and anatomy-aware supervised contrastive learning \cite{chen2023ascon} have expanded the toolkit for robust denoising across diverse datasets \cite{yang2022low} \cite{jung2022patch}. Finally, task-oriented architectures such as TOD-Net \cite{zhang2021task} further improve image quality by emphasizing diagnostically important regions. However, these deep learning approaches face significant limitations in adaptive learning and denoising policy adjustment. Traditional supervised learning methods require extensive datasets and struggle to adapt to varying noise characteristics and imaging conditions in real-time clinical scenarios. Furthermore, these methods often suffer from fixed denoising strategies that cannot dynamically adjust to different noise patterns, leading to suboptimal performance in diverse anatomical contexts and imaging protocols. More importantly, while these methods focus primarily on improving image quality metrics, they often fail to preserve the diagnostically relevant features necessary for downstream clinical tasks such as disease classification and automated diagnosis, limiting their practical clinical utility.

Several methods based on CNNs, GANs, and hybrid or transformer-based networks have been proposed for LDCT denoising \cite{bai2021probabilistic} \cite{tan2022selective} \cite{gu2021adain} \cite{li2021low} \cite{huang2021gan} \cite{de2021gated} \cite{wagner2023benefit} \cite{lee2022unsupervised} \cite{yang2022low} \cite{li2023multi} \cite{li2024unpaired} \cite{zhang2023structure} \cite{bera2021noise} \cite{lei2021strided} \cite{bera2023self} \cite{marcos2022low} \cite{ma2020low}. Although these methods achieve impressive results, they often suffer from limitations such as over-smoothing, instability, and poor generalization across varying noise levels. Moreover, existing approaches lack the ability to learn optimal denoising policies through interactive learning with the imaging environment, which is crucial for handling the complex nature of medical imaging noise and artifacts. However, reinforcement learning (RL) has shown growing potential in LDCT denoising, particularly for adapting to nonstationary noise distributions and fine-tuning filtering parameters \cite{zhang2021r3l} \cite{patwari2022limited} \cite{hu2023reinforcement}. For example, Zhang et al. proposed R3L \cite{zhang2021r3l}, a recurrent reinforcement learning framework that achieves robust image denoising through residual recovery, demonstrating improved generalizability across varying noise levels. Hu et al. provided a comprehensive review of RL applications in medical imaging, highlighting its effectiveness in learning adaptive denoising strategies without relying on large labeled datasets \cite{hu2023reinforcement}. Despite these advances, existing RL-based methods often suffer from training instability and convergence issues, especially with complex medical images, making them less suitable for denoising tasks that require reliable and consistent performance. However, traditional RL algorithms face critical limitations in medical image denoising applications. Conventional policy gradient methods suffer from high variance in gradient estimates, leading to slow convergence and unstable training dynamics. Standard RL approaches also struggle with the continuous action spaces required for fine-grained denoising control and often lack the flexibility to adapt effectively to varying imaging conditions, resulting in suboptimal denoising policies.

\begin{figure}[ht!]
\centering
\includegraphics[scale=0.11]{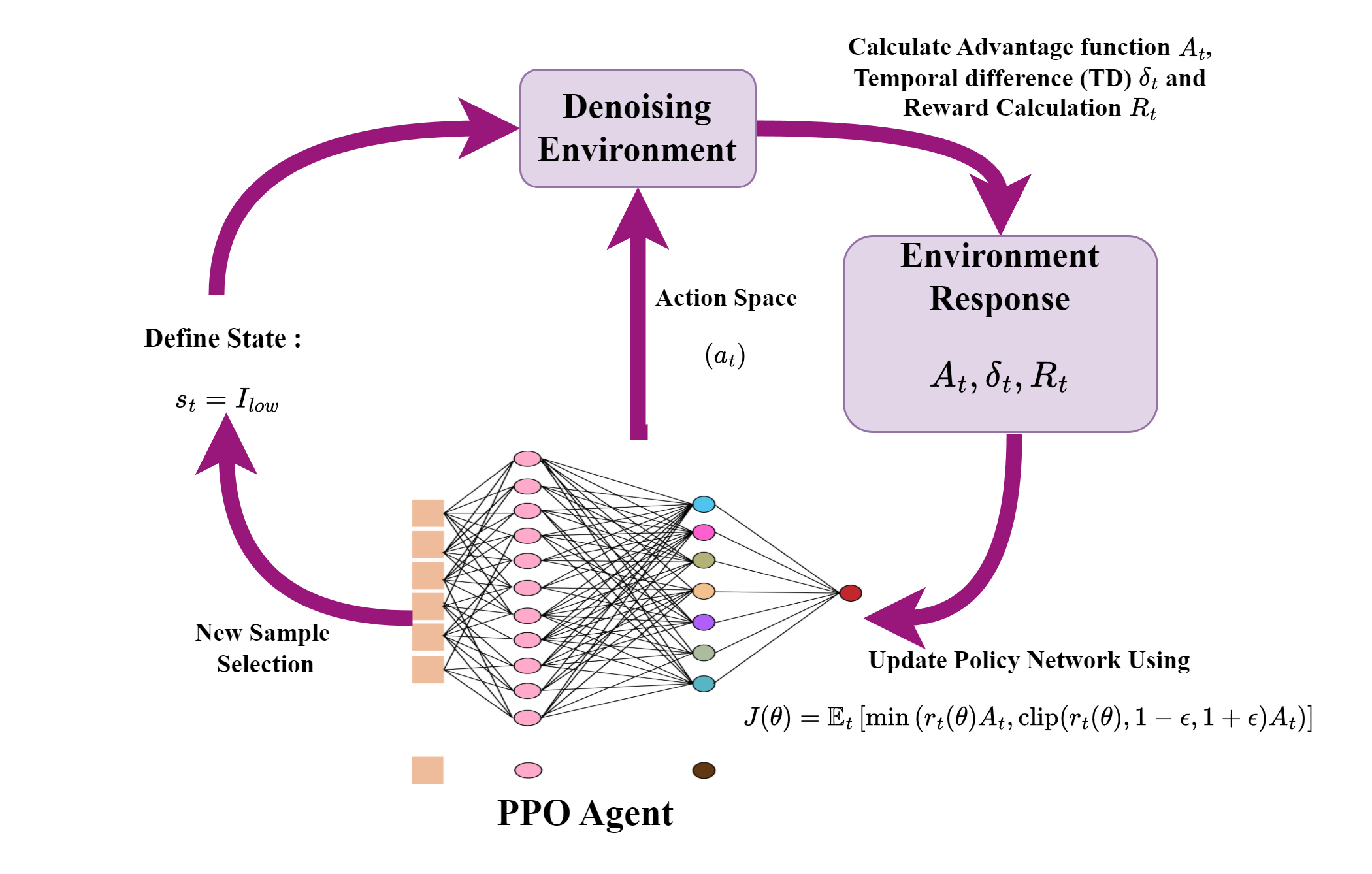}
\caption{The reinforcement learning process of our method}
\label{fig:reinforce}
\end{figure} 

To address these limitations, we propose the following. \\ PPORLD-EDNetLDCT, the first RL-based framework that combines PPO with an encoder-decoder to dynamically adapt denoising policies via real-time image-quality feedback. Our key innovation is using the Proximal Policy Optimization (PPO) as the core learning algorithm, which overcomes the stability and efficiency limitations of conventional RL in medical imaging. PPO's clipped objective function ensures stable learning by preventing unnecessary policy updates while maintaining consistent denoising performance. This approach provides the reliability and computational efficiency needed for clinical deployment and also preserves and enhances diagnostically relevant features, thereby improving performance in downstream clinical applications such as disease classification and automated diagnosis. The ability of PPO to stabilize policy updates and require fewer training episodes enhances its reliability in different noise patterns and imaging protocols, which makes it particularly well suited for clinical denoising tasks. As shown in Figure \ref{fig:reinforce}, the proposed framework uses a custom Gym environment to train the model, which dynamically adapts to different noise levels and imaging conditions. The Gym environment, a standardized framework for RL problems, provides essential components including state representation, action space definition, and reward calculation mechanisms that enable systematic agent-environment interaction for medical image denoising tasks. In this setup, each episode corresponds to the denoising of an LDCT image using an Encoder-Decoder model. The Gym environment provides a standardized interface where the agent interacts with the environment by taking actions, receiving observations, and collecting rewards to learn optimal denoising policies. The reward function is carefully designed based on PSNR and SSIM metrics to reflect image quality and guide the learning process. We use the Proximal Policy Optimization (PPO) algorithm to train the agent, ensuring continuous improvement by maximizing the reward, which directly correlates with better denoising performance. The PPO-based framework learns adaptive denoising strategies that automatically adjust to varying noise characteristics across different imaging protocols and anatomical regions, overcoming the limitations of traditional approaches in clinical LDCT applications.

This comprehensive framework for guiding reinforcement learning with encoder-decoder for low-dose CT (PPORLD-EDNetLDCT) advances LDCT denoising with the potential to significantly improve diagnostic accuracy and patient outcomes by providing high-quality images with reduced noise.

The major contributions of this study are as follows.

\begin{itemize}
    \item This study introduces PPORLD-EDNetLDCT, a reinforcement learning (RL)-based framework for denoising LDCT images that dynamically adapts to varying noise levels and imaging conditions, optimizing image quality without requiring large labeled datasets.

    \item A custom OpenAI Gym-based environment is developed that simulates the denoising process, allowing the agent to interact with noisy images through defined states, actions, and rewards.
    
    \item  An Encoder-Decoder backbone is implemented, incorporating convolutional layers, batch normalization, and transposed convolutions, enhanced using a reinforcement learning strategy to effectively suppress noise while preserving anatomical structures.

    \item The Proximal Policy Optimization (PPO) algorithm is adapted for LDCT denoising, using a custom action space, a reward function based on PSNR and SSIM, and stabilization techniques such as reward clipping and Generalized Advantage Estimation (GAE).

    \item The model is validated on benchmark and unseen datasets, including a COVID-19 LDCT dataset, demonstrating improved generalization and diagnostic performance in a classification setting.
    
\end{itemize}

The rest of the paper is structured as follows. Section \ref{litreview} presents a detailed review of the related literature, Section \ref{method} describes the proposed methodology, Section \ref{results} presents the experimental results, Section \ref{Discussion} offers discussions, and Section \ref{conclusion} concludes the study.

\section{Literature Review}
\label{litreview}
This section provides a review of recent LDCT denoising methods, which we have reorganized into three primary categories, including model-based reconstruction and generative adaptive networks, unsupervised learning, domain adaptation and self-supervised approaches, and attention-based, progressive and hybrid architectures.

\subsection{Model-Based Reconstruction and Generative Adversarial Networks}
Recent developments in LDCT denoising have taken advantage of model-based reconstruction methods and adversarial learning strategies to improve image quality. Model-based approaches focus on explicitly modeling the noise distribution or enhancing structural detail through architectural innovations. For example, Bai et al. \cite{bai2021probabilistic} integrated a shift-invariant neural network into a probabilistic self-learning (PSL) framework, achieving a PSNR of 30.50 and an SSIM of 0.6797 on the AAPM Low-Dose CT Grand Challenge 2016 dataset. Similarly, Li et al. \cite{li2023multi} proposed MSFLNet, a multiscale feature fusion network with attention mechanisms, attaining a PSNR of 33.6490 and an SSIM of 0.9174. These approaches typically achieved moderate PSNR values in the range of 30–33 and SSIM scores up to $\sim 0.92$, demonstrating strong performance in the preservation of anatomical detail.

In contrast, GAN-based methods employ adversarial training to map low-dose images to standard-dose counterparts while preserving perceptual realism. Gu et al. \cite{gu2021adain} introduced an AdaIN-switchable CycleGAN that achieved a PSNR of 30.87 and an SSIM of 0.6605 on multiphase cardiac and chest CT scans. Bera et al. \cite{bera2021noise} improved GAN performance by incorporating a pixel-wise self-attention discriminator, reaching a PSNR of 32.98 and an SSIM of 0.905 in the Mayo Clinic LDCT dataset. Li et al. \cite{li2021low} further advanced this direction with the Denoising SSWGAN framework, which balanced adversarial, perceptual, and structural similarity losses achieving a high SSIM of 0.95649. Other notable contributions include DU-GAN by Huang et al. \cite{huang2021gan}, which operates in the gradient domain (PSNR: 22.31, SSIM: 0.7489), and GRC-GAN by Almeida et al. \cite{de2021gated}, a gated recurrent GAN yielding PSNR 30.2 and SSIM 0.8871 on the LoDoPaB-CT dataset. Also, Mansour et al. \cite{mansour2025aruc} proposed ARUC-GAN, integrating an Attention Residual U-Net within a CycleGAN framework, achieving a PSNR of 34.82 dB and SSIM of 0.85 on the Mayo Clinic dataset while maintaining edge structures with an Edge Keeping Index (EKI) of 0.835.

Specifically, for model‐based reconstruction, Bai et al. \cite{bai2021probabilistic} innovatively integrated a shift-invariant network into a probabilistic self-learning framework, achieving a PSNR of 30.50 and an SSIM of 0.6797 in the AAPM Low-Dose CT Grand Challenge 2016 dataset. Subsequently, Li et al. \cite{li2023multi}’s MSFLNet fused multiscale features with attention‐based modules, obtaining a PSNR of 33.6490 and an SSIM of 0.9174 on the same dataset. Meanwhile, among GAN‐based strategies, Gu et al. \cite{gu2021adain}’s AdaIN-switchable CycleGAN attained a PSNR of 30.87 and an SSIM of 0.6605 on multiphase cardiac and chest CT scans. In contrast, Bera et al. \cite{bera2021noise} introduced a GAN pixel-wise self-attention discriminator that reached a PSNR of 32.98 and an SSIM of 0.905 on the Mayo Clinic LDCT dataset. Similarly, Li et al. \cite{li2021low} introduced the Denoising SSWGAN framework, achieving a PSNR of 32.85 and an SSIM of 0.95649 in the NIH-AAPM Mayo Clinic Grand Challenge. In a more specialized direction, Huang et al. \cite{huang2021gan} proposed DU-GAN, which operates in the gradient domain to produce a PSNR of 22.31 and an SSIM of 0.7489, while Almeida et al. \cite{de2021gated} developed GRC-GAN and obtained a PSNR of 30.2 and an SSIM of 0.8871 on the LoDoPaB-CT dataset.

Many GAN-based models integrate multi-loss designs that combine adversarial, perceptual, sharpness, and SSIM objectives to better preserve fine textures and structural boundaries. For example, SSWGAN’s loss design strategically balances image realism with structural integrity, while DU-GAN’s gradient-domain training enhances edge clarity. Despite their advancements, both model-based and GAN-based methods face limitations such as over-smoothing of fine anatomical details, instability during adversarial training, and a heavy reliance on well-paired training data. Moreover, generalizability across varying noise levels, patient types, and scanner settings remains a persistent challenge, often requiring extensive hyperparameter tuning to stabilize performance.

Additionally, early classical methods like non-local means and block-matching algorithms (e.g., BM3D) laid foundational principles for noise modeling, but often lack the adaptability and perceptual realism achieved by more recent deep learning-based techniques.


\subsection{Unsupervised, Domain Adaptation, and Self-Supervised Methods}
To reduce the dependency on large collections of paired high-dose reference images—which are often difficult, costly, or ethically challenging to obtain—unsupervised and self-supervised learning techniques have emerged as promising alternatives in LDCT denoising. These approaches aim to guide the denoising process without requiring exact input-output image pairs, relying instead on intrinsic image structures, proxy tasks, or domain adaptation techniques.

These methods commonly utilize domain adaptation, deep metric learning, and task-oriented optimization. Notable examples include patchwise deep-metric learning \cite{jung2022patch}, dual-domain self-supervised pipelines \cite{wagner2023benefit}, adversarial domain adaptation from phantom to clinical data \cite{lee2022unsupervised}, task-oriented denoising networks (TOD-Net) \cite{zhang2021task}, strided self-supervision (SN2N) \cite{lei2021strided}, and invertible inter-slice networks \cite{bera2023self}.

For instance, Jung et al. \cite{jung2022patch} proposed a patchwise deep-metric learning strategy, achieving a PSNR of 38.11 and an SSIM of 0.875 on both the NIH-AAPM Challenge and a real-world temporal CT dataset. Wagner et al. \cite{wagner2023benefit} introduced a dual-domain self-supervised pipeline that achieved a PSNR of 41.7 and an SSIM of 0.941 on rebinned helical abdomen CT and cone beam XRM scans. Lee et al. \cite{lee2022unsupervised} utilized adversarial domain adaptation to transfer knowledge from phantom to real image domains, reaching a PSNR of 26.73 and an SSIM of 0.823 on AAPM data. TOD-Net \cite{zhang2021task}, a task-oriented approach designed for lesion-specific enhancement, reported a PSNR of 23.3 and SSIM of 0.767 on the LiTS and KiTS benchmarks. Lei et al. \cite{lei2021strided} introduced the SN2N (Strided Self-Supervision Network), achieving a PSNR of 39.86 and an SSIM of 0.7032 on the Mayo and LIDC-IDRI datasets. Additionally, Lepcha et al. \cite{lepcha2025low} recently introduced a pixel-level nonlocal self-similarity (Pixel-NSS) prior combined with nonlocal means algorithm, achieving a PSNR of 30.3960 and SSIM of 0.8761 on the NIH-AAPM-Mayo dataset while effectively utilizing discrete neighborhood filtering for enhanced computational efficiency. Finally, Bera et al. \cite{bera2023self} developed an invertible inter-slice network, obtaining a PSNR of 31.26 and SSIM of 0.893 on NIH-AAPM and ELCAP datasets.

Furthermore, advanced self-supervised strategies have increasingly focused on diagnostic relevance by integrating supplementary classification networks or invertible architectures. For example, TOD-Net introduces a task-specific loss function that directs denoising toward lesion-sensitive regions critical for downstream analysis. Similarly, invertible networks ensure forward-backward consistency across adjacent slices, thereby enhancing structural integrity. However, while these methods solve the requirement for large paired training data, they still have limitations. Specifically, when noise distributions vary significantly between domains, unsupervised models may converge ineffectively. Additionally, metric learning approaches often struggle to generalize outside the training distribution. Task-oriented strategies, although effective for specific diseases, may unintentionally prioritize localized performance at the cost of overall image quality.

\subsection{Attention-Based and Progressive Hybrid Architectures}
Recently, attention mechanisms and multi-stage hybrid architectures have been widely incorporated into LDCT denoising frameworks to selectively enhance salient regions and iteratively refine image quality. Representative examples include Self-attention Sinogram-to-image Reconstruction (SIST) \cite{yang2022low}; selective kernel-based CycleGANs (SKFCycleGAN) \cite{tan2022selective}; Progressive Wavelet-based Cascaded CNN (PCCNN) \cite{li2024unpaired}; Attention-Guided Context-aware Low-dose CT Reconstruction (AGC-LSRED) \cite{zhang2023structure}; a noise-aware GAN incorporating structural similarity losses \cite{ma2020low}; and FAM-DRL, proposed by Marcos et al. \cite{marcos2022low}, which combines spatial and channel attention with dilated convolutions.

Initially, Yang et al. \cite{yang2022low} proposed the SIST model, which applied self-attention across both projection and image domains, yielding a PSNR of 41.80, SSIM of 0.916, and RMSE of 0.00246 on the Low-Dose CT Image and Projection dataset. Marcos et al. \cite{marcos2022low} (FAM-DRL) reported a PSNR of 40.33 and an SSIM of 0.9102 on piglet, phantom thoracic, and AAPM datasets by incorporating spatial-channel attention and dilated convolutions. Zhang et al. \cite{zhang2023structure} introduced AGC-LSRED using residual attention blocks and global context modeling, achieving PSNR 33.17 and SSIM 0.925 on AAPM-Mayo data. Tan et al. \cite{tan2022selective} developed SKFCycleGAN, leveraging selective kernel fusion and achieving PSNR 41.45, SSIM 0.9535, and RMSE 0.0085 on Mayo and clinical datasets. Li et al. \cite{li2024unpaired} proposed PCCNN with multi-stage wavelet decomposition to retain high-frequency details, yielding PSNR 30.67, SSIM 0.9199, and RMSE 0.029 on the NIH-AAPM dataset. Also, Ma \cite{ma2020low} presented a noise-aware GAN that utilized structural similarity losses and reported PSNR 32.71, SSIM 0.9108, and RMSE 9.49 on the Mayo Challenge dataset. Finally, Zubair et al. \cite{zubair2025novel} presented an attention-guided enhanced U-Net with hybrid edge-preserving structural loss, achieving exceptional performance with a PSNR of 39.6904 and SSIM of 97.67 on the 2016 Low-dose CT AAPM Grand Challenge dataset through the integration of residual blocks and attention gates in up-sampling layers.

Hybrid and progressive schemes—such as PCCNN’s wavelet decomposition and SKFCycleGAN’s kernel-selective mapping reduce noise incrementally while preserving textural fidelity. SIST, by integrating cross-domain attention from projection and image spaces, demonstrates the value of contextual learning for enhancing denoising effectiveness.

However, despite these promising advances, several practical limitations remain. Attention modules often incur high computational overhead, limiting real-time clinical deployment. Progressive models can propagate early-stage reconstruction errors through later layers, and poorly tuned attention modules may amplify noise patterns in extremely low-dose settings. Many methods still over-smooth fine anatomical details, masking clinically significant structures. Moreover, their robustness is often limited across varying noise distributions stemming from differences in scanner types, patient anatomy, and acquisition protocols—resulting in performance inconsistency in real-world applications. Adversarial and hybrid models also demand extensive hyperparameter tuning and can be unstable during training, and their reliance on well-paired high-dose/low-dose datasets limits applicability in data-constrained environments. While reinforcement learning (RL) frameworks have been previously explored for LDCT denoising, existing RL-based approaches often suffer from several critical limitations: unstable training convergence due to high variance in policy gradients, inefficient exploration strategies that lead to suboptimal denoising policies, and difficulties in balancing exploration-exploitation trade-offs during the learning process. Moreover, traditional RL methods frequently struggle with sample efficiency and may require extensive training episodes to achieve satisfactory performance, limiting their practical applicability in clinical settings where computational resources are constrained.

To overcome the limitations of existing denoising techniques and address the shortcomings of previous RL-based methods, we propose a reinforcement learning-based framework that combines Proximal Policy Optimization (PPO) with an enhanced U-Net architecture. The novelty of our approach lies not merely in the use of reinforcement learning, but specifically in the integration of the PPO algorithm within the denoising process. PPO addresses the key limitations of conventional RL methods by providing stable policy updates through clipped objective functions, ensuring more reliable convergence and improved sample efficiency. Our methodological contribution focuses on the specific design and implementation of the PPO-based adaptive denoising strategy, which dynamically adapts to varying noise levels while maintaining training stability. This method models denoising as a policy learning problem, where the agent optimizes actions based on reward signals derived from PSNR and SSIM metrics. The framework ensures consistent and high-fidelity reconstruction of low-dose CT images across diverse imaging conditions, without relying on large paired training datasets.

\section{Methodology}
\label{method}
This section outlines the datasets, the model architecture, including the Proximal Policy Optimization (PPO) framework, and the reinforcement-based training approach utilized in our study to denoise low-dose CT scans.

\begin{figure*}[ht!]
\centering
\includegraphics[scale=0.45]{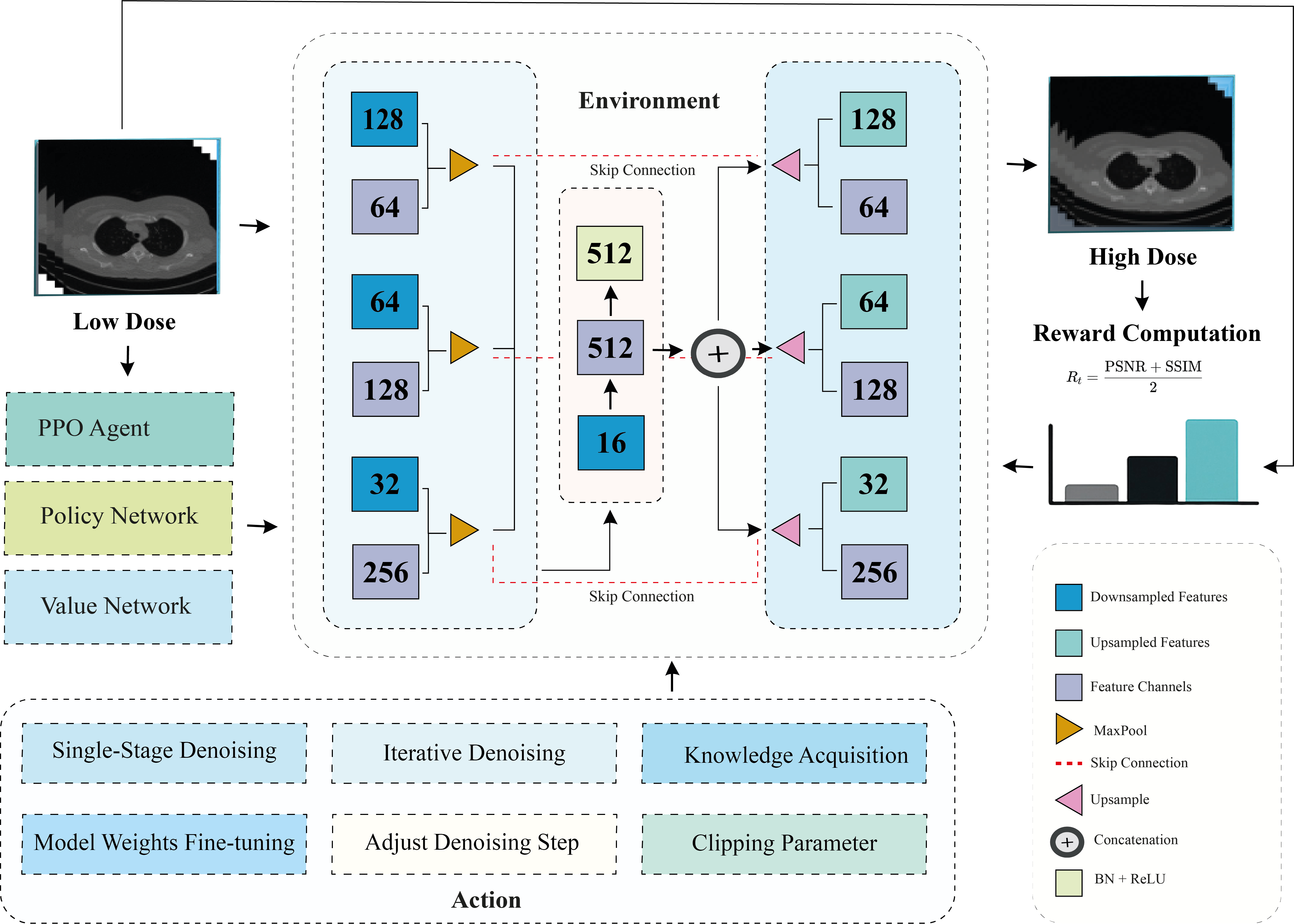}
\caption{\textbf{Overview of PPORLD-EDNetLDCT}. The proposed framework utilizes a PPO (Proximal Policy Optimization) agent which interacts with an environment based on an Encoder-Decoder. The low-dose CT image is input into the encoder, where features are progressively downsampled. Then these features are subsequently upsampled and concatenated via skip connections to reconstruct a denoised output. A reward signal is computed based on PSNR and SSIM metrics comparing the output with the high-dose ground truth. This reward is used to guide actions such as model fine-tuning, iterative denoising, adjusting denoising steps, and other strategies to optimize the denoising process iteratively, which helps the model to adaptively improve image quality.}
\label{fig:overall}
\end{figure*}
\subsection{Proposed Model: PPORLD-EDNetLDCT}
This study proposes an RL based encoder-decoder model with skip connection modules. While the architectural components follow a standard encoder-decoder design, our primary novelty lies in integrating this network with a reinforcement learning framework using Proximal Policy Optimization (PPO). This combination enables adaptive denoising actions based on real-time feedback, rather than relying on fixed supervised losses alone. Figure~\ref{fig:overall} illustrates the complete architecture of the proposed denoising framework, PPORLD-EDNetLDCT, which integrates a reinforcement learning strategy with an encoder-decoder-based network. The framework highlights the role of a Proximal Policy Optimization (PPO) agent that intelligently guides and optimizes the denoising process through adaptive decision-making.
\subsubsection{Encoder}
Proposed  PPORLD-EDNetLDCT model initiates the process with an encoder that decomposes the input image into a hierarchical feature representation using 64 convolutional filters of size 3 $\times$ 3. Here, the input image $X \in \mathbb{R}^{H \times W \times C}$ represents the low-dose CT image where $H$ and $W$ are the spatial dimensions and $C$ is the number of channels. It captures essential low-level details such as edges and textures while maintaining the original spatial dimensions. This is followed by a second encoding layer. This layer reduces the spatial resolution by half and doubles the number of feature mappings to 128 with a stripped convolution. The strided convolution operation effectively performs downsampling while simultaneously learning hierarchical representations, where the spatial dimensions become $\frac{H}{2} \times \frac{W}{2}$ with 128 feature channels. It allows the model to focus on more relative patterns in the image. The third encoding layer continues to lower spatial dimensions while raising the feature maps' depth to 256 to further extraction of high-level features. At this stage, the feature maps have spatial dimensions of $\frac{H}{4} \times \frac{W}{4}$ with 256 channels, capturing increasingly abstract representations of the input CT image. Each layer in the encoding path is enhanced with ReLU activation to introduce non-linearity and batch normalization to stabilize the training process. In the encoder, each layer extracts hierarchical features, as represented in Equation \eqref{eq:eq_1} :
\begin{equation}
    F_{l+1} = \sigma\left( \text{BN}\left( W_l F_l + b_l \right) \right) \quad l = 1, 2, \dots, L
    \label{eq:eq_1}
\end{equation}

where \( F_l \) is the input feature map at layer \( l \), \( W_l \) is the convolutional weights, \( b_l \) is the bias term, \( \ast \) denotes the convolution operation, \( \text{BN}(\cdot) \) represents batch normalization, and \( \sigma(\cdot) \) is the activation function, in this case its ReLU.

\subsubsection{Bottleneck Layer}
The encoder and decoder are connected through the bottleneck layer, which operates at the lowest spatial resolution of the architecture. This layer expands the feature space to 512 channels, enabling the model to encapsulate the most abstract and high-dimensional representations of the input. The bottleneck layer processes features at spatial resolution $\frac{H}{4} \times \frac{W}{4}$ and serves as the information compression point where the most semantically rich representations are formed. The bottleneck acts as a bridge between the encoding and decoding paths, extracting high-dimensional features at the lowest spatial resolution to capture the core structure of the image, discard irrelevant details, and prepare the features for reconstruction in the decoder.This is represented in Equation \eqref{eq:eq_2}:

\begin{equation}
\label{eq:eq_2}
    F_b = \sigma\left( \text{BN}\left( W_b \ast F_{\text{enc}} + b_b \right) \right) \tag{2}
\end{equation}

where \( F_b \) is the bottleneck feature map, \( b_b \) is bias addition and \( F_{\text{enc}} \) is the output feature map from the encoder.

After the bottleneck, three corresponding decoder layers systematically reconstruct the image: the first decoder layer upsamples from 512 to 256 channels, the second from 256 to 128 channels, and the third from 128 to 64 channels, with each layer doubling spatial dimensions and integrating corresponding encoder features via skip connections.
\subsubsection{Decoder}
The decoding path mirrors the structure of the encoder. Its primary purpose is to reconstruct the denoised image by gradually restoring the spatial resolution. The first decoding layer begins this process by upsampling the features from the bottleneck layer. This is achieved through transposed convolutions, which double the spatial dimensions of the feature maps. Transposed convolution performs learned upsampling where the stride parameter controls the upsampling factor, effectively reversing the downsampling operation performed in the encoder. Additionally, skip connections linked to this decoding layer with the corresponding encoding layer help the model to reuse high-level features extracted during encoding. This process is repeated in the second and third decoding layers. Additionally, these layers refine the reconstruction by integrating mid-level and low-level features, ensuring that the output retains both the global structure and fine-grained features of the original image. The decoder reconstructs the image by upsampling the feature maps and integrating skip connections. Equation \eqref{eq:eq_3} contains the mathematical representation of the decoding layer : 

\begin{equation}
\label{eq:eq_3}
    F_{l-1} = \sigma\left( \text{BN}\left( W_l^{\top} \ast F_l + b_l \right) \right) + F_{\text{skip}} \tag{3}
\end{equation}

where \( W_l^{\top} \) represents the transposed convolutional filters used for upsampling, \( F_{\text{skip}} \) are the feature maps from the encoder introduced via skip connections, and \( F_l \) and \( F_{l-1} \) are the feature maps at consecutive layers of the decoder.

\subsubsection{Final Output Layer}
The encoder's feature maps are incorporated by the decoder using skip connections. These links ensure that the reconstruction procedure restores fine-grained spatial features that may have been lost during the encoder downsampling. The denoised image is produced at the end of the decoder by passing the final feature map through a convolutional layer. Equation \eqref{eq:eq_4} mathematically represents the output convolutional layer : 

\begin{equation}
\label{eq:eq_4}
    \hat{X} = W_{\text{out}} \ast F_d + b_{\text{out}} \tag{4}
\end{equation}

where \( \hat{X} \) is the denoised output image, and \( W_{\text{out}}, b_{\text{out}} \) are the parameters of the output convolutional layer.

\subsubsection{Skip Connections}
An important part of this architecture is the skip connections. This mechanism directly transfers feature maps from the encoding layers to their corresponding decoding layers. Skip connections address the loss of spatial details during downsampling, which is one of the primary challenges in image reconstruction. The skip connections improve the model's ability to precisely rebuild structural information by reinstating these details in the decoding process. This trait is especially crucial for medical imaging, where maintaining the accuracy of the anatomical details is crucial. Skip connections implement element-wise addition between encoder and decoder feature maps at corresponding spatial resolutions, ensuring that both low-level spatial information and high-level semantic information are preserved throughout the reconstruction process.  The skip connection mechanism is represented by Equation \eqref{eq:eq_5} :

\begin{equation}
\label{eq:eq_5}
    F_{\text{dec}} = F_{\text{dec}} + F_{\text{enc}} \tag{5}
\end{equation}

where \( F_{\text{dec}} \) is the decoder feature map, and \( F_{\text{enc}} \) is the corresponding encoder feature map.

\subsubsection{Proximal Policy Optimization (PPO)}

Proximal Policy Optimization (PPO) is a state-of-the-art reinforcement learning (RL) algorithm used in this pipeline to optimize the denoising decision-making process \cite{schulman2017proximal}. In our framework, it is adapted for the LDCT denoising task. The agent is implemented as a Multi-Layer Perceptron (MLP) and is trained to interact with a custom environment, where each state represents a noisy CT image and the Encoder-Decoder model acts as the environment being manipulated. Specifically, it consisted of three fully connected layers, an input layer matching the flattened state representation, two hidden layers with 256 and 128 neurons respectively using ReLU activation, and an output layer with a softmax activation producing probabilities over the discrete action space. This design balanced expressive capacity with computational efficiency, ensuring stable training within our low-dose CT denoising environment. To enable PPOs to make context-aware denoising decisions, we design a domain-specific action space comprising operations like applying the Encoder-Decoder, reapplying it, fine-tuning model weights, or skipping processing. The reward function is carefully constructed using a weighted combination of Peak Signal-to-Noise Ratio (PSNR) and Structural Similarity Index Measure (SSIM) and is clipped to improve training stability. 

We also integrate generalized advantage estimation (GAE) to reduce the variance in policy updates and ensure smoother learning. These targeted enhancements make PPO highly effective in guiding the denoising process in a dynamic, image quality-driven manner, extending its typical usage from discrete control tasks to continuous high-dimensional medical image optimization. PPO iteratively learns by adjusting its policy to maximize these rewards to ensure that the model continuously improves denoising performance. The PPO agent operates within the environment, where each state \( s_t \) corresponds to a low-dose CT image that requires denoising. The policy function \( \pi_{\theta} (a_t \mid s_t) \) maps the input state \( s_t \) to an action \( a_t \), determining the next processing step. The policy function is parameterized as in Equation \eqref{eq:eq_6} :

\begin{equation}
\label{eq:eq_6}
    \pi_{\theta} (a_t \mid s_t) = P(a_t \mid s_t; \theta) \tag{6}
\end{equation}

where \( \theta \) represents the trainable parameters of the MLP.

The PPO agent interacts with the environment by selecting discrete actions that determine how the model (Encoder-Decoder) is applied to low-dose CT images. We design the action space to include various denoising maneuvers, such as applying the encoder-decoder network or adjusting its parameters At each time step, the policy network (MLP) receives the current image state and chooses one of five possible actions:

\begin{enumerate}
    \item \textbf{Apply Encoder-Decoder Once}: Applies the (Encoder-Decoder) model once for basic noise reduction.
    
    \item \textbf{Apply Multiple Times}: Applies the model 2-3 times iteratively for stronger denoising, where the output of one application becomes the input for the next iteration.
    
    \item \textbf{Fine-tune Model Parameters}: Fine-tunes the model's trainable parameters dynamically. This action allows the agent to fine-tune the denoiser's weights on-the-fly by one gradient descent step (using the current image's MSE loss). This helps the agent explore improvements to the base model for particularly noisy inputs. Although this makes the environment non-stationary, we mitigate instability via reward clipping and PPO's robust policy updates.
    
    \item \textbf{Skip Processing}: Skips the denoising step to preserve the current state.
    
    \item \textbf{Adjust PPO Learning Parameters}: Adjusts PPO's own learning parameters for improved stability. This action dynamically modifies the PPO learning rate and exploration parameters (such as entropy coefficient and value function coefficient) within safe bounds based on recent reward variance. When action 3 (fine-tuning) causes training instability due to the non-stationary environment, this action helps stabilize training by reducing the learning rate temporarily and increasing the value function weight to improve policy-value alignment.
\end{enumerate}

The chosen action is then executed within the environment, where the processed image is evaluated using PSNR and SSIM metrics, and a reward is computed based on the image quality improvement. We chose to average PSNR (in dB) and SSIM for reward. This gives strong weight to pixel-level accuracy while still incorporating structural similarity though SSIM's contribution is smaller in magnitude. We found this balance effective in guiding the agent toward solutions that prioritize reconstruction fidelity while maintaining perceptual quality. Equation \eqref{eq:eq_7} and \eqref{eq:eq_8} represent the PSNR and SSIM Equations. Equation \eqref{eq:eq_9} shows the reward \( R_t \) calculation.

\begin{equation}
\label{eq:eq_7}
    \text{PSNR} = 10 \log_{10} \left( \frac{\text{MAX}_I^2}{\text{MSE}} \right) \tag{7}
\end{equation}
where $\text{MAX}_I$ is the maximum possible pixel value of the image
\begin{equation}
\label{eq:eq_8}
\resizebox{\columnwidth}{!}{$
    \text{SSIM}(I_{\text{denoised}}, I_{\text{groundtruth}}) =
    \frac{(2 \mu_x \mu_y + C_1)(2 \sigma_{xy} + C_2)}
         {(\mu_x^2 + \mu_y^2 + C_1)(\sigma_x^2 + \sigma_y^2 + C_2)}
$} \tag{8}
\end{equation}

where $\mu_x$ and $\mu_y$ represent the mean intensities of the denoised and ground truth images, respectively; $\sigma_x^2$ and $\sigma_y^2$ are their variances; $\sigma_{xy}$ denotes the covariance between them; and $C_1$, $C_2$ are small constants to avoid instability when the denominator is close to zero.

\begin{equation}
\label{eq:eq_9}
\resizebox{\columnwidth}{!}{
    $R_t = \frac{\text{PSNR}(I_{\text{denoised}}, I_{\text{groundtruth}}) + \text{SSIM}(I_{\text{denoised}}, I_{\text{groundtruth}})}{2}$
}
\tag{9}
\end{equation}

The agent learns the optimal denoising technique in a number of episodes by using this reward to alter its policy during the PPO optimization process. PPO continuously improves its decision-making process by using this reinforcement learning loop, learning to apply denoising adaptively while maintaining anatomical structures in medical images. PPO computes an advantage function to determine whether a selected action led to a better than expected outcome. To refine the advantage estimation, the Generalized Advantage Estimation (GAE) \cite{schulman2015high} is used. GAE balances the bias-variance trade-off in estimating the advantage function by introducing a decay parameter λ, which allows the model to incorporate short-term and long-term reward information more effectively. This results in smoother learning and more reliable policy gradients during PPO training. Equation \eqref{eq:eq_10} presents the advantage estimation function.

\begin{equation}
\label{eq:eq_10}
    A_t = \sum_{l=0}^{\infty} (\gamma \lambda)^l \delta_{t+l} \tag{10}
\end{equation}

where \( A_t \) is the advantage estimate at time step \( t \), \( \gamma \) is the discount factor, controlling how much future rewards are considered, \( \lambda \) is the GAE decay parameter, balancing bias and variance in advantage estimation, and \( \delta_{t+l} \) represents the temporal difference (TD) residual at time \( t+l \). Temporal difference \( \delta_t \) is explained through Equation \eqref{eq:eq_11}:

\begin{equation}
\label{eq:eq_11}
    \delta_t = R_t + \gamma V(s_{t+1}) - V(s_t) \tag{11}
\end{equation}

where \( R_t \) is the immediate reward at time \( t \), \( V(s_t) \) is the value function estimate for state \( s_t \), and \( V(s_{t+1}) \) is the estimated value of the next state \( s_{t+1} \).

However, without constraints, excessively high rewards could lead to overfitting or encourage over-smoothing, which may degrade anatomical structures in medical images. As shown in Equation \eqref{eq:eq_12}, to mitigate this, the reward is capped, enforcing a range between 0 and 100.

\begin{equation}
\label{eq:eq_12}
    R_t = \max(0, \min(R_t, 100)) \tag{12}
\end{equation}

This capping ensures that negative or misleading penalties do not discourage learning while also preventing excessively large updates that could destabilize the PPO agent. Additionally, NaN values in PSNR or SSIM are replaced with zero, ensuring numerical stability. By constraining the reward range, the PPO model gradually optimizes its denoising strategy without over-prioritizing extreme cases, leading to a more stable and generalized policy that preserves essential image details while reducing noise.

After computing the reward, PPO updates its policy network (MLP) using a clipped surrogate objective function to ensure controlled and stable learning. Equation \eqref{eq:eq_13} represents the surrogate objective function \cite{schulman2017proximal}.

\begin{equation}
\label{eq:eq_13}
    J(\theta) = \mathbb{E}_t \left[ \min \left( r_t(\theta) A_t, \text{clip}(r_t(\theta), 1-\epsilon, 1+\epsilon) A_t \right) \right] \tag{13}
\end{equation}

where \( r_t(\theta) \) is the probability ratio between the updated and previous policies, \( A_t \) is the advantage function, which determines whether the selected action was better than expected, and \( \epsilon \) is the clipping parameter, preventing excessive updates. By combining reinforcement learning-based policy updates (PPO) with traditional supervised learning (MSE loss), the model balances pixel-level accuracy and perceptual quality improvements.
We set the PPO discount factor $\gamma=0.99$, GAE $\lambda=0.95$, clipping $\epsilon=0.2$, and used an AdamW optimizer with initial learning rate $\eta=5 \times 10^{-5}$ for the encoder-decoder model and $\eta_{PPO}=1 \times 10^{-4}$ for the PPO policy network.

\subsubsection{Reinforcement-Based Training}
The training process is built upon a reinforcement learning framework, where the model is integrated into a custom environment defined using the Gymnasium framework \cite{towers2024gymnasium}. The environment manages data loading, interaction with the model, and reward computation. The training process begins with the initialization of the PPO policy network and denoising model. At each episode, a low-dose CT image is provided as input, and the agent selects an action from a discrete set. The action set includes applying model once, multiple times, fine-tuning model parameters, skipping denoising, or adjusting PPO hyperparameters. The selected action is executed, and the denoised image is generated and evaluated against the high-dose ground truth using Peak Signal-to-Noise Ratio (PSNR) and Structural Similarity Index (SSIM). The computed reward, defined as the average of PSNR and SSIM, is capped within a stable range (0-100) to ensure robust learning.

PPO updates the policy network using a clipped surrogate objective to prevent drastic updates. The Mean Squared Error (MSE) loss between the denoised and high-dose images serves as an auxiliary supervised learning signal that is backpropagated through the encoder-decoder denoising model to fine-tune its parameters, while the PPO loss is used separately to update the policy network that controls action selection. The PPO model is trained for multiple episodes, iteratively improving its policy to maximize the expected reward. Algorithm \ref{alg:PPO_CT_Denoising} represents the overall process of training. In each episode (Algorithm 1 lines 5–34), the environment is reset with a new low-dose image (state $s_0$). Then for each time step $t$ (lines 7–33), the agent observes the current image state and samples an action $a_t \sim \pi_\theta(s_t)$, executes it (lines 11–26 for each action case), obtains a reward $R_t$ (line 27) and next state $s_{t+1}$, and stores the transition. The dual optimization approach ensures that while the policy network learns optimal action selection strategies through reinforcement learning, the denoising model parameters are simultaneously refined through supervised learning when fine-tuning actions are selected.

\begin{algorithm} [ht!]
\footnotesize
\caption{PPO-Based Training for Low-Dose CT Image Denoising}
\label{alg:PPO_CT_Denoising}
\begin{algorithmic}[1]
    \State \textbf{Input:} \textit{DenoiseEnv} (RL environment), Low-dose CT images $I_{low}$, High-dose CT images $I_{high}$, PPO policy $\pi_{\theta}$, Encoder-Decoder denoising model, $N_{episodes}$ (max episodes), $N_{steps}$ (max steps per episode).
    
    \vspace{0.05cm} 
    \State \textbf{Output:} Trained model for optimal denoising.
    
    \vspace{0.1cm} 
    
    \State Initialize PPO policy $\pi_{\theta}$ with random weights and model.
    \vspace{0.05cm} 
    \State Set up \textit{DenoiseEnv}.
    
    \vspace{0.05cm} 
    \For{each episode $e$ from $1$ to $N_{episodes}$}
        \State Reset environment, select $I_{low}$ and $I_{high}$.
        \vspace{0.05cm} 
        \For{each step $t$ from $1$ to $N_{steps}$}
            \State Set state $s_t = I_{low}$.
            \State Sample action $a_t \sim \pi_{\theta}(s_t)$.
            \State \textbf{Perform action:}
            \indent
            \State \textbf{case} Apply model once:
            \indent
                \State $I_{denoised} = \text{Encoder-Decoder}(I_{low})$

            \State \textbf{case} Apply model multiple times:
            \indent
                \State $I_{denoised} = I_{low}$
                \For{$k = 1$ to $K$}
                    \State $I_{denoised} = \text{Encoder-Decoder}(I_{denoised})$
                \EndFor

            \State \textbf{case} Fine-tune model:
            \indent
                \State Compute loss $\mathcal{L} = \| \text{Encoder-Decoder}(I_{low}) - I_{high} \|^2$
                \State Update $\omega \leftarrow \omega - \eta \nabla_\omega \mathcal{L}$
                \State $I_{denoised} = \text{Encoder-Decoder}(I_{low})$

            \State \textbf{case} Skip denoising:
            \indent
                \State $I_{denoised} = I_{low}$

            \State \textbf{case} Adjust PPO hyperparameters:
            \indent
                \State Modify PPO parameters
                \State $I_{denoised} = \text{Encoder-Decoder}(I_{low})$

            \State Compute reward $R_t = (PSNR + SSIM)/2$, clip to $(0,100)$.
            \State Store $(s_t, a_t, R_t, s_{t+1})$ in PPO memory.
            \State Optimize PPO:
            \indent
                \State Compute $L_{MSE} = MSE(I_{denoised}, I_{high})$.
                \State Compute PPO loss using clipped objective.
                \State Update $\pi_{\theta}$ using AdamW optimizer.
            
        \EndFor
        \vspace{0.05cm} 
    \EndFor
    \vspace{0.05cm} 
\end{algorithmic}
\end{algorithm}
During the training phase, the model utilizes dynamic actions, such as Adjust PPO Learning Parameters and Fine-tune Model Parameters, to optimize the performance of the encoder-decoder network. These actions allow the model to adapt the learning rate and refine its parameters based on feedback from the reward function and loss function. This adaptation ensures stable convergence and improving denoising performance over time.

However, it is important to note that these actions are only valid during training. They are not applicable during inference, as the model has already been trained and the parameters are fixed. During inference, the model performs a straightforward denoising pass, using the learned, fixed parameters to process new data. As a result, inference time remains computationally efficient, unaffected by the dynamic adjustments made during training.

\section{Results}
\label{results}
\subsection{Dataset}
\textbf{Low Dose CT Image and Projection Dataset \cite{moen2021low}:} This dataset, sourced from The Cancer Imaging Archive (TCIA), contains clinically acquired chest CT scans alongside corresponding simulated low-dose projection data. The simulations were generated using a validated noise-insertion method across two CT scanner types under routine clinical protocols. In this study, only the reconstructed low-dose chest CT images were used for training—projection data were not utilized. The dataset consists of 100 scans. These were split into 80\% for training and 20\% for testing, with no patient overlap between sets. Given the reinforcement learning setup, each scan was reused across multiple training episodes, allowing the agent to iteratively refine its denoising policy. No explicit data augmentation was applied; however, all CT slices were used directly. The images were normalized to [0, 1] intensity range and resized to a uniform resolution of 512×512 pixels.

\textbf{COVID-19 Low-Dose and Ultra-Low-Dose CT scans \cite{afshar2022human}:} This dataset was employed solely for validation purposes. It comprises 160 chest CT scans, including 56 normal cases and 104 COVID-19-positive cases. Among the COVID-19 cases, 36.5\% were confirmed via RT-PCR, while the remaining diagnoses were established by consensus among three radiologists. The dataset contains both low-dose and ultra-low-dose CT images, which were treated uniformly during evaluation, without separate handling. This diversity in case types and dose levels provided a robust testbed for assessing the generalization capability of the proposed method.

\textbf{NIH-AAPM-Mayo Clinic Low Dose CT Challenge \cite{mccollough2017low}:} This dataset has 30 contrast-enhanced abdominal CT scans acquired at the Mayo Clinic as part of the 2016 NIH–AAPM–Mayo Clinic Low-Dose CT Grand Challenge. Each case was scanned in the portal venous phase on a Siemens SOMATOM Flash system, with full-dose acquisitions at 120 kV/200 mAs and simulated quarter-dose data at 120 kV/50 mAs. As this is a widely used dataset for low dose denoising task, we utilized this dataset to compare our methodology with a wide range of studies.

\begin{figure}[ht!]
\centering
\includegraphics[scale = 0.2]{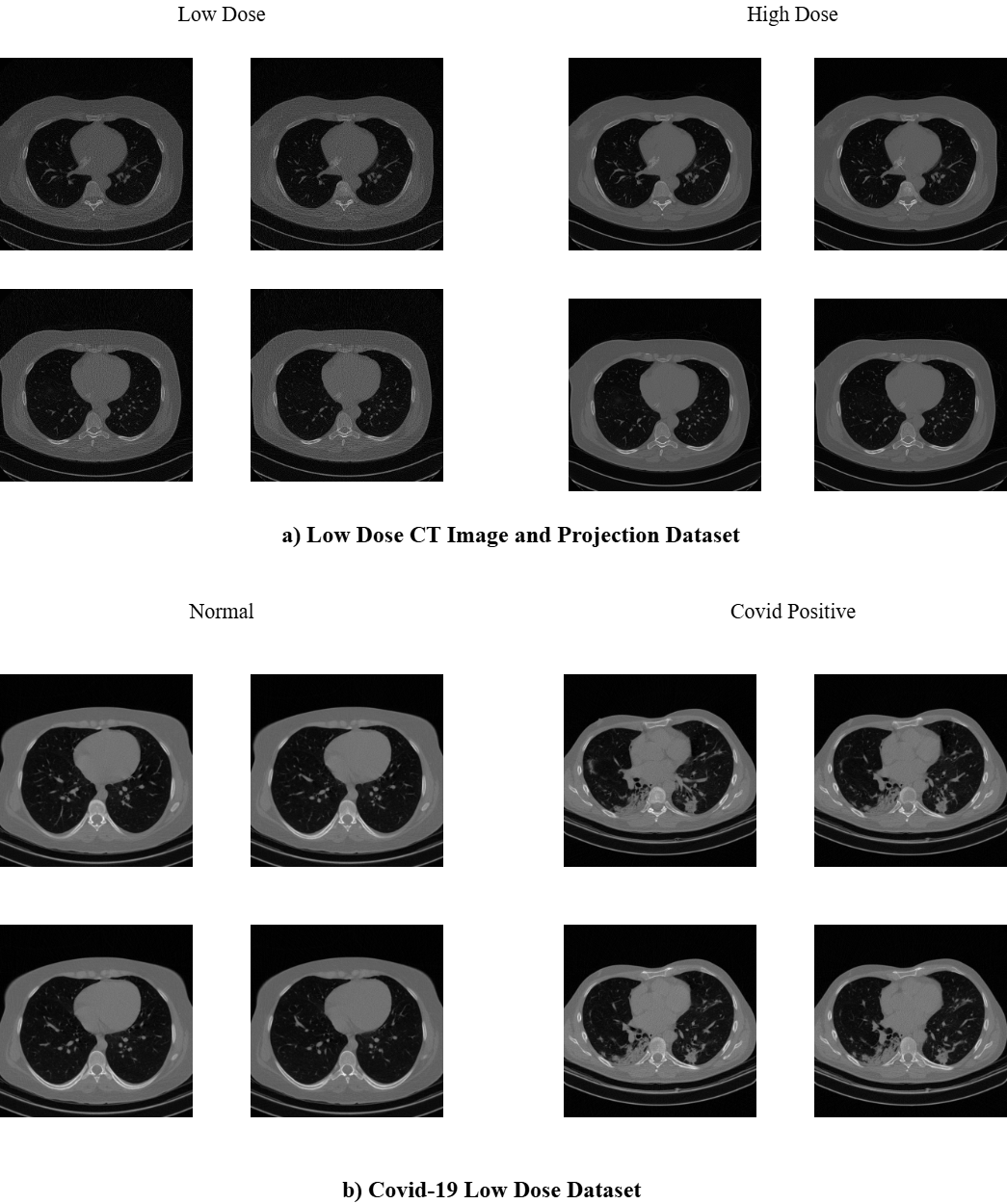} 
\caption{Sample images from Low-Dose CT Image and Projection Dataset and the COVID-19 Low-Dose Dataset}
\label{fig:fig_1}
\end{figure}

Table \ref{tab:datasets} summarizes the usage of the datasets and Figure \ref{fig:fig_1} shows sample images from the datasets.

\begin{table}[H]
\label{tab:datasets}
\centering
\caption{Datasets used in this study}
\scriptsize
\begin{tabular}{|l|c|c|l|}
\hline
\textbf{Dataset} & \textbf{Scans} & \textbf{Use}  \ \\
\hline
Low Dose CT (TCIA) \cite{moen2021low} & 100 (chest) & Training   \\
\hline
COVID-19 CT \cite{afshar2022human} & 160 & Validation \\
\hline
LDCT Grand Challenge \cite{mccollough2017low} & 30(abdomen) & Comparison \\
\hline
\end{tabular}
\label{tab:datasets}
\end{table}

\subsection{Training Result}
\begin{figure}
\centering
\includegraphics[scale=.2]{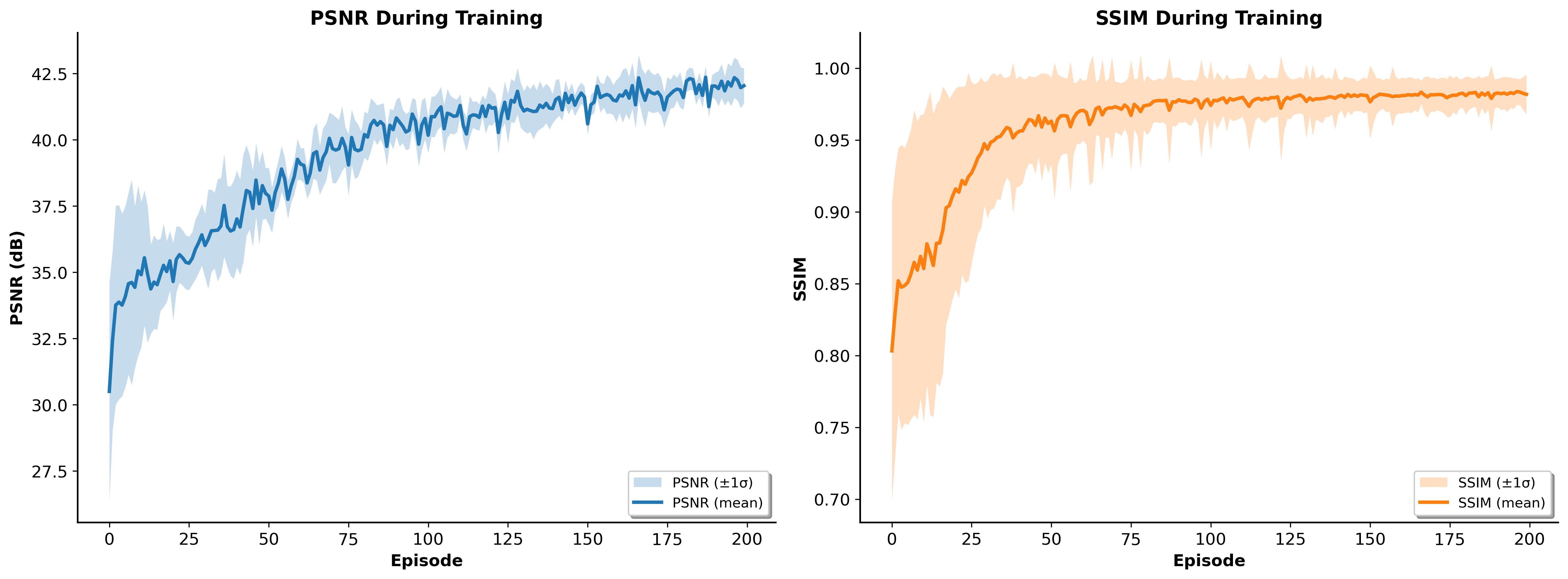} 
\caption{PSNR and SSIM progression over training epochs on Low Dose CT Image and Projection Dataset}
\label{fig:fig_2}
\end{figure}

The Model is trained using the Low Dose CT Image and Projection Dataset. Figure \ref{fig:fig_2} demonstrates the performance progression of our model over 200 training epochs in a reinforcement environment, with plots for PSNR (Peak Signal-to-Noise Ratio) and SSIM (Structural Similarity Index Measure). The PSNR curve (left plot) exhibits a rapid increase during the initial epochs, reflecting the model's ability to quickly learn and improve pixel-level accuracy in denoising low-dose CT images. This is followed by a gradual but consistent improvement as the model converges toward optimal performance. Similarly, the SSIM curve (right plot) shows significant early gains, indicating the model's effectiveness in preserving structural and perceptual details.The shaded regions around the PSNR and SSIM curves represent the ±1 standard deviation error margins, capturing the variability across validation samples during training. In the PSNR plot, the error margin is initially wider, reflecting greater performance fluctuation in early episodes, but it progressively narrows. This indicates increased stability and consistency as training advances. Similarly, in the SSIM plot, the error band tightens significantly after the early phase, suggesting that the model achieves reliable structural similarity across different samples with minimal variance by the end of training. These error margins help quantify the confidence and robustness of the model’s performance. Together, these results highlight the model's capacity to effectively enhance image quality while maintaining both pixel-level fidelity and structural integrity throughout the training process.

Moreover, we further trained and evaluated our model on the NIH–AAPM–Mayo Clinic Low-Dose CT Grand Challenge dataset and observed similar performance trends to those seen on the LDCT Image and Projection dataset. Both the PSNR and SSIM curves showed rapid early improvements followed by stable convergence with progressively narrowing error margins, underscoring the stability of the reinforcement learning process. Quantitatively, the model achieved a PSNR of 41.52, a SSIM of 0.9723, and a RMSE of 0.0051 on this dataset. These results are consistent with the earlier findings, strengthening our claim that the proposed method generalizes effectively across different CT regions. Together, these insights confirm the robustness of our approach in enhancing image quality while preserving structural fidelity and diagnostic information across multiple benchmarks.

\begin{table}[h]
    \centering
    \caption{Comparison with State-of-the-Art Denoising Methods on Low Dose CT Image and Projection Dataset}
    \fontsize{7pt}{8pt}\selectfont 
    \renewcommand{\arraystretch}{1.4}
    \begin{tabular}{@{}lccc@{}} 
        \toprule
        \textbf{Method} & \textbf{PSNR ($\uparrow$)} & \textbf{SSIM ($\uparrow$)} & \textbf{RMSE ($\downarrow$)} \\
        \midrule
        PWLS-ULTRA \cite{zheng2018pwls} & 38.98 & 0.902 & 0.00337 \\
        RED-CNN \cite{tan2022selective} & 40.31 & 0.908 & 0.00289 \\
        DP-ResNet \cite{yin2019domain} & 40.92 & 0.914 & 0.00269 \\
        Attention-Guided U-Net \cite{zubair2025novel} & 39.69 & 0.9767 & 0.0107 \\
        Pixel-NSS \cite{lepcha2025low} & 30.40 & 0.8761 & 0.0302 \\
        ARUC-GAN \cite{mansour2025aruc} & 34.82 & 0.850 & 0.0180 \\
        SIST \cite{yang2022low} & 41.80 & 0.916 & 0.00246 \\
        \textbf{PPORLD-EDNetLDCT} & \textbf{41.87} & \textbf{0.9814} & \textbf{0.00236} \\
        \bottomrule
    \end{tabular}
    \label{tab:denoising_comparison}
\end{table}

\begin{table}[!ht]
    \small
    \centering
    \caption{Comparison of results on COVID-19 Low-Dose and Ultra-Low-Dose CT scans dataset before and after applying Reinforcement Learning (Encoder-Decoder = ED)}
    \fontsize{7pt}{8pt}\selectfont 
    \renewcommand{\arraystretch}{1.1}
    \begin{tabular*}{\linewidth}{@{\extracolsep{\fill}}lcc}
        \toprule
        \textbf{Metric} & \textbf{With ED} & \textbf{With PPORLD-EDNetLDCT} \\
        \midrule
        Accuracy & 0.90 & 0.94 \\
        F1-Score & 0.88 & 0.92 \\
        Precision & 0.87 & 0.93 \\
        Recall & 0.89 & 0.91 \\
        Specificity & 0.88 & 0.92 \\
        AUC-ROC & 0.91 & 0.95 \\
        AUC-PR & 0.85 & 0.91 \\
        MCC & 0.78 & 0.85 \\
        Balanced Accuracy & 0.89 & 0.93 \\
        False Positive Rate & 0.12 & 0.08 \\
        False Negative Rate & 0.11 & 0.09 \\
        \bottomrule
    \end{tabular*}
    \label{tab:rl_comparison}
\end{table}

\newcolumntype{L}[1]{>{\raggedright\arraybackslash}p{#1}} 
\newcolumntype{X}{>{\raggedright\arraybackslash}X}         

\begin{table*}[ht!]
    \centering
    \caption{Comparison with Existing Literature}
    \fontsize{8pt}{8pt}\selectfont 
    \begin{tabularx}{\textwidth}{c L{6.5cm} L{5cm} c c c}
        \toprule
        \textbf{Paper} & \textbf{Dataset} & \textbf{Model} & \textbf{PSNR} & \textbf{SSIM} & \textbf{RMSE} \\ 
        \midrule
        \cite{bai2021probabilistic} & AAPM Low-Dose CT Grand Challenge 2016 & PSL & 30.50 & 0.6797 & - \\ 
        \cite{gu2021adain} & Multiphase Cardiac CT scans, Chest CT Scans & AdaIN-switchable cycleGAN & 30.87 & 0.6605 & - \\
        \cite{bera2021noise} & Mayo Clinic Low-dose CT image database & Self-attention and pixel-wise GANs & 32.98 & 0.905 & 9.69 \\
        \cite{zhang2021task} & LiTS, KiTS & TOD-Net & 23.3 & 0.767 & 28.5 \\
        \cite{yang2022low} & Low-Dose CT Image and Projection Dataset, Simulated Dataset & SIST & 41.80 & 0.916 & 0.00246 \\
        \cite{jung2022patch} & NIH-AAPM-Mayo Clinic Low Dose CT Challenge, real-world temporal CT scan dataset & Patch-wise deep metric learning & 38.11 & 0.875 & - \\
        \cite{wagner2023benefit} & Rebinned helical abdomen CT scans, X-ray microscope (XRM) scans & Dual-domain and self-supervised CT denoising pipeline & 41.7 & 0.941 & - \\
        \cite{li2023multi} & NIH AAPM Mayo Clinic Low-Dose CT Challenge dataset & MSFLNet & 33.65 & 0.9174 & - \\
        \cite{lei2021strided} & Mayo LDCT, LIDC IDRI & SN2N & 39.86 & 0.7032 & - \\
        \cite{li2021low} & NIH-AAPM-Mayo Clinic LDCT Grand Challenge, lung CT dataset, real piglet CT dataset & Denoising SSWGAN & 32.85 & 0.95649 & - \\
        \cite{de2021gated} & LoDoPaB-CT Dataset & GRC-GAN & 30.2 & 0.8871 & 0.11 \\
        \cite{bera2023self} & NIH-AAPM-Mayo Clinic LDCT Grand Challenge, ELCAP Public Lung Image Database & Invertible Network Exploiting Inter Slice Congruence & 31.26 & 0.893 & - \\
        \cite{huang2021gan} & NIH-AAPM-Mayo Clinic LDCT Grand Challenge, Real-World Dataset & DU-GAN & 22.31 & 0.7489 & 0.0802 \\
        \cite{marcos2022low} & Piglet and Phantom Thoracic datasets, AAPM grand challenge database & FAM-DRL & 40.33 & 0.9102 & - \\
        \cite{ma2020low} & NIH-AAPM-Mayo Clinic LDCT Grand Challenge & Noise learning GAN & 32.71 & 0.9108 & 9.49 \\
        \cite{marcos2024generative} & CT simulation of a deceased piglet by Yi and Babyn & RED-GAN & 37.59 & 0.9102 & - \\
        \cite{li2024unpaired} & NIH-AAPM-Mayo Clinic LDCT Grand Challenge, Low Dose CT Image and Projection Data & PCCNN & 30.67 & 0.9199 & 0.029 \\
        \cite{tan2022selective} & Mayo dataset, Clinical dataset & SKFCycleGAN & 41.45 & 0.9535 & 0.0085 \\
        \cite{zhang2023structure} & AAPM-Mayo dataset & AGC-LSRED & 33.17 & 0.925 & 9.02 \\
        \cite{lee2022unsupervised} & NIH-AAPM-Mayo Clinic LDCT Grand Challenge & Adversarial Domain Adaptation & 26.73 & 0.823 & - \\
        \midrule
        \textbf{Our Study} & \textbf{Low Dose CT Image and Projection Dataset} & \textbf{PPORLD-EDNetLDCT} & \textbf{41.87} & \textbf{0.9814} & \textbf{0.00236} \\
        \textbf{Our Study} & \textbf{NIH-AAPM-Mayo Clinic Low Dose CT Challenge Dataset} & \textbf{PPORLD-EDNetLDCT} & \textbf{41.52} & \textbf{0.9723} & \textbf{0.0051 }\\
        \bottomrule
    \end{tabularx}
    \label{tab:comparison_literature}
\end{table*}

\begin{figure*}[!ht]
\centering
\includegraphics[width=\linewidth]{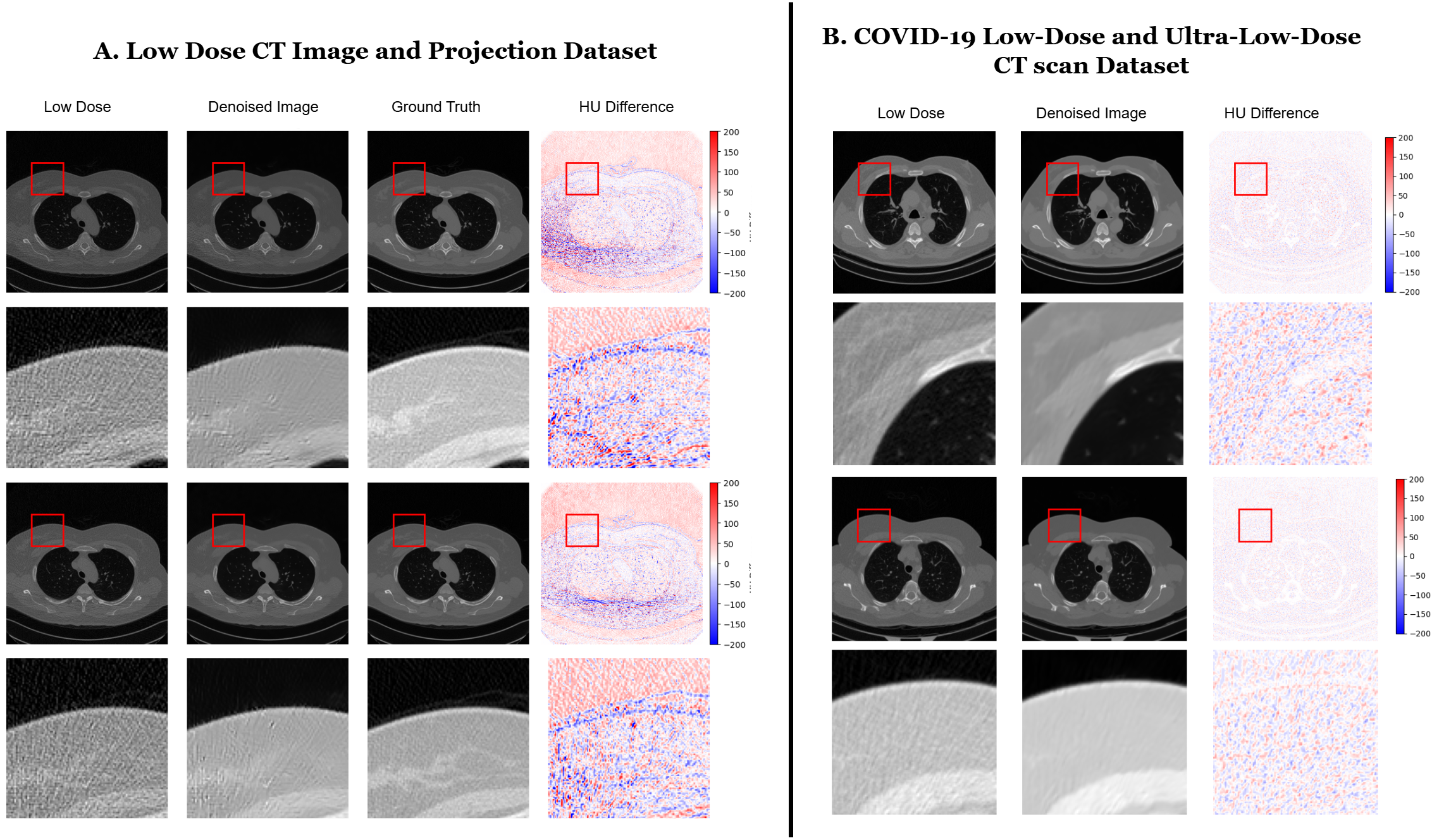}
\caption{Comparative analysis of low-dose and denoised CT images with HU difference visualizations for quality assessment, where (A) is from the TCIA dataset and (B) is from the COVID-19 dataset.}
\label{fig:fig_3}
\end{figure*}

\subsection{Comparison With Other Methods}
Table \ref{tab:denoising_comparison} presents a comprehensive comparison between our method and state-of-the-art denoising approaches on the Low Dose CT Image and Projection Dataset. The evaluation includes traditional reconstruction methods, deep learning-based models, and recent advanced techniques.

The results demonstrate that iterative techniques like PWLS-ULTRA \cite{zheng2018pwls} achieve moderate performance, while deep learning approaches show superior results. For example, PWLS-ULTRA \cite{zheng2018pwls} achieves a PSNR of 38.98 and an SSIM of 0.902, while RED-CNN \cite{tan2022selective} reaches a PSNR of 40.31 and SSIM of 0.908. DP-ResNet \cite{yin2019domain} attains a PSNR of 40.92 with SSIM 0.914, and SIST \cite{yang2022low} records a PSNR of 41.80 and SSIM of 0.916, representing previous state-of-the-art levels. Recent methods demonstrate diverse performance levels. ARUC-GAN \cite{mansour2025aruc} obtains a PSNR of 34.82 and an SSIM of 0.85, while the Attention-Guided U-Net \cite{zubair2025novel} delivers improved results with a PSNR of 39.69 and an SSIM of 0.9767. In contrast, Pixel-NSS \cite{lepcha2025low} achieves a lower PSNR of 30.40 and an SSIM of 0.8761.

Regarding RMSE, our PPORLD-EDNetLDCT achieves 0.00236, outperforming all compared methods and confirming superior noise suppression and detail preservation.

Our PPORLD-EDNetLDCT achieves the best overall performance with a PSNR of 41.87 and an SSIM of 0.9814, outperforming all compared methods while maintaining superior structural preservation. The quantitative results confirm our model's robustness across varying noise levels and anatomical regions.

\subsection{Validation With Unseen Dataset}
To evaluate the robustness and generalizability of the model, validation experiments were carried out using the COVID-19 Low-Dose and Ultra-Low-Dose CT scans dataset. These experiments were designed to compare the performance of the model with and without the integration of the PPORLD-EDNetLDCT framework.  For denoising the images, two pipelines were used. At first, our proposed model was trained with normal settings, and then the same model was trained with reinforcement learning. In both scenarios, a pre-trained ResNet-50 backbone was used to ensure a consistent and robust feature extraction pipeline for classification. The ResNet-50 classifier was fine-tuned on the denoised images rather than trained from scratch, with the final fully connected layer adapted for COVID-19 vs. normal classification. This approach leverages pre-trained ImageNet features while adapting to the specific characteristics of the denoised CT images. The model was trained and validated for 30 epochs and a comprehensive set of performance metrics was used to assess its effectiveness. Table \ref{tab:rl_comparison} compares the performance metrics of the pipelines with and without reinforcement learning. All classification performance metrics improved when using RL-denoised images as depicted in Table \ref{tab:rl_comparison}. Notably, accuracy rose from 90\% to 94\%, and the AUC-ROC from 0.91 to 0.95, indicating that our denoiser better enhanced the features needed for distinguishing COVID-19 from normal cases. This underscores that our RL framework's noise reduction did not distort pathology, but actually made it easier for the classifier to detect. The comparative analysis underscores the impact of our method in enhancing the model's performance. By improving both pixel-level accuracy and structural similarity during the denoising process, the PPORLD-EDNetLDCT framework contributes to superior classification outcomes.

\subsection{Comparison With Existing Studies}
Table~\ref{tab:comparison_literature} presents a structured summary of recent state-of-the-art approaches in low-dose CT (LDCT) denoising, highlighting their reported performance in terms of PSNR, SSIM and RMSE across various datasets. The primary goal of including this table is to contextualize our proposed PPORLD-EDNetLDCT model within the broader landscape of LDCT denoising research. By combining our results with those from established literature, we aim to demonstrate the relative strengths of our method in terms of reconstruction quality and noise suppression. Our method outperforms GAN-based approaches such as denoising SSWGAN \cite{li2021low} and SKFCycleGAN \cite{tan2022selective} by effectively preserving high-frequency textures and anatomical structures while reducing noise. The proposed reinforcement learning framework enables the model to dynamically adapt to varying noise levels, unlike traditional supervised models such as MSFLNet \cite{li2023multi}, which require large amounts of paired data. Compared to domain adaptation methods such as the dual-domain self-supervised pipeline \cite{wagner2023benefit}, our approach achieves better PSNR and RMSE by optimizing both structural fidelity and perceptual quality through the reward-based policy. Furthermore, the inclusion of an adaptive policy (via PPO) in our framework achieves noise suppression without over-smoothing, a balance that even advanced methods like SN2N \cite{lei2021strided} or FAM-DRL \cite{marcos2022low} struggle with. The table provides a valuable reference point for understanding the progress of LDCT denoising research and underscores the potential of our approach to achieve state-of-the-art performance. In particular, our method achieved a PSNR of 41.87, SSIM of 0.9814, and RMSE of 0.00236 on the AAPM Low Dose CT dataset, and a PSNR of 41.52, SSIM of 0.9723, and RMSE of 0.0051 on the NIH-AAPM-Mayo Clinic dataset Low Dose CT Challenge dataset. These results highlight the robustness and effectiveness of our method in denoising LDCT images while preserving fine structural details and diagnostic information across different datasets.

\newcolumntype{L}[1]{>{\raggedright\arraybackslash}p{#1}}
\newcolumntype{X}{>{\raggedright\arraybackslash}X}

\begin{table*}[!ht]
    \scriptsize
    \centering
    \caption{Ablation Study of the PPORLD-EDNetLDCT Framework on LDCT Denoising with Statistical Significance and Performance Testing}
    \begin{tabularx}{\textwidth}{c L{5.5cm} c c c c c c}
        \toprule
        \textbf{ID} & \textbf{Configuration} & \textbf{PSNR (↑)} & \textbf{SSIM (↑)} & \textbf{RMSE (↓)} & \textbf{Wilcoxon p} & \textbf{Train (s/epoch)} & \textbf{Infer (ms)} \\
        \midrule
        Full Model & Encoder-Decoder + PPO + GAE + reward clipping + custom Gym & \textbf{41.87} & \textbf{0.9814} & \textbf{0.00236} & -- & \textbf{320} & \textbf{46} \\
        A1 & EDNet-Single (Encoder-Decoder only, supervised MSE loss) & 29.52 & 0.8427 & 0.00302 & 0.00001 & 180 & 44 \\
        A2 & EDNet-Multi (fixed 3-pass denoising, no PPO) & 31.08 & 0.8891 & 0.00291 & 0.00002 & 240 & 47 \\
        A3 & PPO without reward clipping & 25.13 & 0.9602 & 0.00264 & 0.00000 & 310 & 46 \\
        A4 & PPO without Generalized Advantage Estimation (GAE) & 30.41 & 0.9638 & 0.00259 & 0.00010 & 305 & 46 \\
        A5 & PPO with fixed action set (no dynamic adaptation) & 32.25 & 0.9586 & 0.00273 & 0.00120 & 295 & 45 \\
        A6 & PPO-ApplyOnly (policy limited to single/multi apply actions) & 36.92 & 0.9694 & 0.00251 & 0.00800 & 285 & 45 \\
        A7 & Encoder-Decoder without skip connections & 36.79 & 0.9281 & 0.00342 & 0.00500 & 200 & 44 \\
        A8 & Reward function using only PSNR & 38.28 & 0.9490 & 0.00268 & 0.01200 & 315 & 46 \\
        A9 & Reward function using only SSIM & 33.02 & 0.9631 & 0.00275 & 0.00750 & 290 & 45 \\
        \bottomrule
    \end{tabularx}
    \label{tab:ablation_study}
\end{table*}

\subsection{Ablation Study}
Table \ref{tab:ablation_study} highlights the contribution of each component in the PPORLD-EDNetLDCT framework, combining performance metrics with statistical validation and computational efficiency. To ensure that the observed differences were not due to random variation across test cases, we performed significance testing on per-image PSNR, SSIM, and RMSE values. For each ablation, the difference between the full model and the ablated variant was computed on the same set of images, and the Wilcoxon signed-rank test was applied. This non-parametric test evaluates whether the median of paired differences differs significantly from zero without assuming normality. In all comparisons, the Wilcoxon test yielded $p$-values below 0.05, with most below 0.01, confirming that the improvements achieved by PPORLD-EDNetLDCT over the ablations are statistically robust.

Removing PPO (A1) causes the PSNR to drop sharply from 41.87 to 29.52 (Wilcoxon $p=0.00001$), demonstrating that reinforcement learning is essential for adaptive decision-making. Although A1 is faster to train, the severe loss in reconstruction quality highlights the importance of the reward-driven adaptive approach. Adding multiple fixed passes without PPO (A2) improves PSNR slightly to 31.08 but remains far below the full model and statistically significant ($p=0.00002$), while requiring longer training time.

Disabling reward clipping (A3) reduces PSNR further to 25.13, reflecting severe instability from unbounded rewards. This was the weakest configuration overall, with Wilcoxon $p<0.00001$. Excluding GAE (A4) leads to 30.41 dB PSNR due to high-variance policy updates ($p=0.00010$), while restricting the agent to a fixed action set (A5) limits adaptability and increases RMSE ($p=0.00120$). Both train slightly faster than the full model, but at the cost of clear quality degradation. Allowing only simple “apply once/multiple” actions (A6) yields moderate performance (36.92 dB, $p=0.00800$), confirming that sequential scheduling learned by PPO is beneficial but still inferior to the full framework.

Removing skip connections (A7) raises RMSE to 0.00342 and reduces structural fidelity, with statistical tests confirming significance ($p=0.00500$). Training also becomes faster, showing that the skip connections increase cost but are necessary for detail preservation. Reward design is equally critical, using only PSNR (A8) or only SSIM (A9) weakens performance, with the SSIM-only reward producing oversmoothing and a sharp PSNR drop to 33.02 ($p=0.00750$). Training times for these reward-only variants (315 and 290 s/epoch) remain close to the full model, reinforcing that reward shaping primarily impacts quality rather than efficiency.

Notably, inference times remain stable across all variants, showing that the proposed framework achieves significant accuracy and robustness gains without sacrificing inference efficiency. The main computational load occurs in training where reinforcement learning utilizes iterative parameter tuning.

\subsection{Qualitative Evaluation of the PPORLD-EDNetLDCT Model}
Figure \ref{fig:fig_3} illustrates the effect of denoising on example images: (a) from the TCIA dataset, (b) from the COVID-19 dataset. The difference images (right side of each) show mostly gray with speckles of blue/red, indicating noise removal with minimal structural change. The red areas correspond to slight increases in Hounsfield Units after denoising – these should be scrutinized to ensure they don't distort pathology. It helps to visualize how much the low-dose CT scan changes after denoising by comparing it pixel by pixel with the denoised scan. On the left side example of Low Dose CT Image and Projection Dataset is presented. On the right side example of COVID-19 Low-Dose and Ultra-Low-Dose CT scans dataset is shown. It employs a coolwarm colormap, with blue areas denoting a drop in intensity and red areas showing an increase in Hounsfield Unit (HU) values (indicating that those areas were more intense after denoising). Gray or neutral areas indicate little to no change, indicating that such structures were maintained during the denoising process. A decent denoising method should ideally eliminate noise without sacrificing significant anatomical information. Nonetheless, if the difference image exhibits structured patterns, it may indicate that the denoising procedure has brought about some systematic alterations that could affect medical diagnosis.

\section{Discussion}
\label{Discussion}
Earlier research in LDCT denoising has achieved notable progress through various deep learning approaches. Model-based reconstruction methods like PSL \cite{bai2021probabilistic} and MSFLNet \cite{li2023multi} achieved moderate PSNR values but often produced over-smoothed images. GAN-based approaches including AdaIN-switchable CycleGAN \cite{gu2021adain}, SSWGAN \cite{li2021low}, and DU-GAN \cite{huang2021gan} preserved texture information better but suffered from training instability and mode collapse issues. Attention-based methods such as SIST \cite{yang2022low}, SKFCycleGAN \cite{tan2022selective}, and AGC-LSRED \cite{zhang2023structure} improved feature representation but required extensive computational resources. However, these existing approaches rely on fixed denoising strategies and require extensive datasets, which limits their effectiveness across diverse anatomical structures and imaging protocols.

Our proposed denoising framework (PPORLD-EDNetLDCT) for low-dose CT (LDCT), based on reinforcement learning (RL), offers an innovative and highly adaptive method to enhance medical images. Unlike conventional supervised deep learning approaches that depend on fixed models with pre-set weights, our RL-based method continually evolves and refines its denoising strategy by interacting with data. Using Proximal Policy Optimization (PPO), the model dynamically adjusts its denoising process in response to real-time feedback from image quality assessments. This allows it to adapt to different noise levels, specific patient traits, and various imaging conditions without the need for extensive manual adjustments. While previous RL approaches like R3L \cite{zhang2021r3l} demonstrated promising results in adaptive denoising, they faced challenges with training stability and consistent convergence. Our study addresses these limitations by introducing PPO's clipped objective function and constrained policy update mechanism, ensuring stable learning dynamics and consistent performance improvements that existing RL methods could not achieve.  Although the encoder-decoder backbone used in this study follows a conventional structure with convolutional layers, batch normalization, and transposed convolutions, its capabilities are significantly enhanced through the integration of reinforcement learning. By incorporating Proximal Policy Optimization (PPO), the model transitions from a static denoising network to a dynamic, policy-driven system capable of adapting its denoising strategy in real time based on feedback from image quality metrics. This synergy between RL and encoder-decoder architecture allows the model to better preserve fine anatomical structures while suppressing noise, particularly in challenging LDCT scenarios. Thus, innovation lies not in architectural design alone, but in how reinforcement learning fundamentally increases and governs the behavior of the network during inference.

Another major strength lies in the use of Proximal Policy Optimization (PPO) \cite{schulman2017proximal}. In our framework, PPO is customized with a domain-specific action space for low-dose CT denoising, including options such as applying the Encoder-Decoder once or multiple times, fine-tuning its parameters, skipping denoising or adjusting PPO hyperparameters. These actions are executed within a custom environment, where each state represents a noisy image and the reward is computed based on PSNR and SSIM to reflect image quality improvements. The reward is cut to maintain stability, and the NaN values are explicitly handled to prevent training disruptions. This represents a significant advancement over traditional policy gradient methods that exhibit unstable convergence, require extensive hyperparameter tuning, and struggle with learning efficiently from limited data. It adds adaptive exploration, which directs the model towards the best reconstruction techniques that preserve fine details while lowering noise, where the typical supervised model learns only from pixel-wise loss functions such as MSE. This combination of reinforcement learning and supervised learning enables the model to generalize better, making PPO an essential component of this image denoising pipeline.  

Further advantage of our framework is its ability to self-regulate denoising intensity. Unlike existing methods that apply uniform denoising strategies regardless of image characteristics, our PPO-based approach learns to select appropriate denoising actions based on individual image properties. This adaptive capability addresses a critical gap in current literature where methods fail to handle the non-stationary nature of medical imaging noise. While the model has been primarily validated on chest CT scans, its design suggests strong potential for performance across other anatomical regions, such as the abdomen and brain. The model’s dynamic denoising approach, driven by reinforcement learning, is inherently adaptable to various CT acquisition protocols. This flexibility enables the model to adjust to differing noise distributions and imaging artifacts.  Future validation across multiple anatomical regions and CT protocols is necessary, but the model’s robust architecture and training approach strongly suggest it will generalize well across a wide range of clinical conditions.

In contrast, our PPO-based model assesses each image individually, selecting the most suitable denoising action from a structured action space that includes single-pass denoising, multi-pass enhancement, real-time parameter adjustments, and PPO policy refinements. This allows it to achieve a balance between noise suppression and the preservation of fine anatomical structures, making it particularly useful for medical diagnostics. Additionally, our approach does not require extensive manually labeled datasets, unlike many GAN-based and supervised learning models. Instead, the RL model optimizes its denoising strategy based on real-time feedback using Peak Signal-to-Noise Ratio (PSNR) and Structural Similarity Index (SSIM) as reward metrics, ensuring that noise suppression does not compromise diagnostic quality. This directly addresses the dataset dependency limitations of existing supervised approaches, while the stable training characteristics of PPO ensure reliable convergence that previous RL methods could not achieve consistently. While the Proximal Policy Optimization (PPO) and dynamic action strategies contribute to a higher computational cost during the training phase, they do not significantly impact inference time. The model's foundation is a encoder-decoder structure, which ensures that inference can be performed efficiently with moderate computational resources. The main computational load occurs in training, where reinforcement learning necessitates iterative parameter tuning and optimization. However, once the model is trained, its deployment in clinical practice is expected to be computationally feasible, with minimal latency in real-time applications.

\subsection{Limitations}
However, certain challenges remain in our approach, particularly regarding generalization across different scanners and imaging protocols \cite{keenan2022challenges}. While our model is designed to dynamically adapt to varying noise levels and imaging conditions, its performance is still influenced by the characteristics of the training data. The model demonstrates strong results on the TCIA and COVID-19 LDCT data sets, but a larger validation is necessary to confirm its robustness across a wider spectrum of imaging protocols and scanner types. For example, evaluation on datasets acquired using different CT protocols (such as brain CT and dual-energy CT) and from various vendors (such as GE and Philips) would help account for scanner-specific artifacts, reconstruction algorithms, and dose-modulation strategies. These factors can introduce variability in image quality and noise characteristics, which can affect the consistency of the model in diverse clinical settings. Future exploration of domain adaptation techniques or few-shot fine-tuning could help further improve generalization.

Another consideration is the potential for over-smoothing in certain cases \cite{xu2024uncovering}.Although the model is optimized for PSNR and SSIM, which are widely accepted image quality metrics, these measures primarily assess pixel-wise similarity and may not fully reflect the preservation of fine diagnostic details, such as small pulmonary nodules. Although the integration of reinforcement learning reduces the risk of loss of structural information, some degree of blurring may still occur, potentially masking subtle features such as early-stage tumors or small vascular anomalies.

Although our approach significantly reduces the dependency on manually labeled datasets, it currently relies on surrogate quality metrics such as PSNR and SSIM. These may not always align perfectly with clinical evaluation standards \cite{breger2025study}. However, these limitations highlight important opportunities for future research, including the integration of clinically informed metrics and validation across multicenter datasets to further improve the practical applicability and reliability of the model in clinical workflows.

\subsection{Future Works}
To address these limitations and improve the overall performance of our denoising framework, several future directions can be explored. A promising avenue is hybrid approaches, where reinforcement learning is combined with self-supervised learning \cite{li2022does} or transformer-based architectures to improve denoising efficiency while maintaining adaptability. This would allow the model to learn from unlabeled datasets, further reducing dependency on paired training data. Modern architecture such as attention mechanisms or transformer-based encoders can be also employed in future rather than traditional encoder-decoder. Thus, when combined with our reinforcement learning framework, it may further improve structure preservation and adaptivity in complex LDCT scenarios .  To ensure that model does not lose any important detail like nodules, incorporating more clinically guided loss functions can be a solution in future studies. Specifically, reward functions can be designed using different nodule preserving loss functions such as radiomics-based evaluation or lesion-preserving losses. Incorporating attention-based mechanisms to the model can be another addition to keep critical structures intact. To enhance generalization, future work can focus on domain adaptation techniques. This ensures that our model performs robustly across different scanner types, clinical settings, and patient populations. This could involve training the model on multi-center datasets to improve its adaptability to diverse imaging conditions. Additionally, incorporating task-aware reward functions that prioritize diagnostically significant features, rather than relying solely on traditional image similarity metrics, could lead to further improvements in clinical applicability. 

\section{Conclusion}
\label{conclusion}

The proposed reinforcement learning-based denoising framework offers a highly adaptive, scalable, and data-efficient alternative to conventional LDCT noise reduction techniques. The proposed reinforcement learning-based denoising framework achieves strong performance with a PSNR of 41.87, SSIM of 0.9814, and RMSE of 0.00236 in LDCT Image and Projection dataset, demonstrating its ability to reduce noise while preserving important anatomical structures. By integrating reinforcement learning with deep learning, it dynamically refines denoising strategies in real-time, ensuring an optimal balance between noise suppression and anatomical detail preservation. Unlike traditional methods that rely on static parameters or extensive labeled datasets, this approach leverages policy optimization and reward-driven learning to enhance generalization across diverse imaging conditions. 

With further advancements in domain adaptation, hybrid learning techniques, and clinically driven optimization, for example, adversarial domain adaptation or normalization techniques for denoising images with different noise distributions, this framework has the potential to revolutionize medical imaging by enabling safer, low-dose CT scans without compromising diagnostic accuracy, ultimately contributing to improved patient outcomes and broader accessibility to high-quality radiological assessments. The proposed framework holds significant promise for real-world implementation in clinical settings, including hospitals, diagnostic imaging centers, and telemedicine platforms. By providing real-time denoising support, it can enhance image clarity, facilitate accurate diagnoses, and reduce the dependency on high-dose radiation exposure. However, this model’s generalizability remains limited by the diversity of training data and imaging protocols. Future work should focus on validating the model across datasets from different scanners and clinical settings to ensure robustness and generalizability, enabling deployment in clinical environments where clearer, diagnostically useful images can support radiologists while minimizing patient radiation exposure.


\section*{Declarations}
\textbf{Conflict of Interests:} On behalf of all authors, the corresponding author states that there is no conflict of interest.\\
\textbf{Funding:} No external funding is available for this research.\\
\textbf{Data Availability Statement:} This study conducts all the experiments on 2 publicly available datasets: Low Dose CT Image and Projection Dataset \cite{moen2021low} and COVID-19 Low-Dose and Ultra-Low-Dose CT scans \cite{afshar2022human}.\\
\textbf{Ethics Approval and Consent to Participate}. Not applicable \\
\textbf{Informed Consents:} All the authors have read the manuscript fully and contributed to this research. The authors also provide consent for peer review and publication of this study.\\
\textbf{CRediT authorship contribution statement:}

\textbf{Debopom Sutradhar}: Conceptualization, Methodology, Writing – original draft; 
\textbf{Ripon Kumar Debnath}: Resources, Literature review, Writing – original draft; 
\textbf{Mohaimenul Azam Khan Raiaan}: Conceptualization, Writing – original draft, Validation, Formal analysis, Writing – review \& editing; 
\textbf{Reem E. Mohamed}: Validation, Formal analysis, Writing – review \& editing; 
\textbf{Yan Zhang}: Validation, Formal analysis, Writing – review \& editing; 
\textbf{Sami Azam}: Validation, Formal analysis, Writing – review \& editing, Supervision, Project administration.


\end{document}